\documentclass{article}

\PassOptionsToPackage{numbers, compress}{natbib}

\usepackage[final]{neurips_2020}

\usepackage[utf8]{inputenc} 
\usepackage[T1]{fontenc}    
\usepackage{hyperref}       
\usepackage{url}            
\usepackage{booktabs}       
\usepackage{amsfonts}       
\usepackage{nicefrac}       
\usepackage{microtype}      

\title{MSR-Net: Multi-Scale Relighting Network for One-to-One Relighting}

\author{%
  Sourya Dipta Das\thanks{Equal contribution} \\
  Jadavpur University, India\\
  \texttt{dipta.juetce@gmail.com} \\
   \And
   Nisarg A. Shah$^*$ \\
   IIT Jodhpur, India \\
   \texttt{shah.2@iitj.ac.in} \\
   \AND
   Saikat Dutta \\
   IIT Madras, India \\
   \texttt{saikat@smail.iitm.ac.in} \\
}

\usepackage{graphicx}
\begin{document}

\maketitle

\begin{abstract}
  Deep image relighting allows photo enhancement by illumination-specific retouching without human effort and so it is getting much interest lately. Most of the existing popular methods available for relighting are run-time intensive and memory inefficient. Keeping these issues in mind, we propose  the  use  of  Stacked Deep  Multi-Scale  Hierarchical  Network, which  aggregates features from each image at different scales. Our solution is differentiable and robust for translating image illumination setting from input image to target image. Additionally, we have also shown that using a multi-step training  approach to this problem with two different loss functions can significantly boost performance and can achieve a high quality reconstruction of a relighted image. 
\end{abstract}

\section{Introduction}
Most of the professional tools have limited ability to manipulate light for small fluctuations in intensity or color. Designers are required to spend endless hours to get the appropriate settings for their images with limited options of enhancements being possible. In contrast, this can be solved on a large scale by using deep learning-based methods with the availability of higher variability. This has massive applicability in research as well as in practice. There are multiple instances where relighting based methods would be helpful for artistic enhancement. Some of them include its usage in applications like editing of photos, manipulation of shadows, and variation of the color temperature of the image. Image relighting can also be used for data augmentation to increase the amount of training data, which will make the model robust to changes in color temperature or position of light source. It can also be used for domain adaptation, in which input images will be transformed from different illumination settings into a unique set of illumination settings that the particular network was trained upon. Here, the problem description is to transfer one fixed set of illumination settings from an input image taken at a scene to another fixed set of illumination settings as target image at same scene. In this paper, we  propose  the  use  of  both Deep  Multi-Scale  Hierarchical  Network and Stacked Deep Multi-Scale  Hierarchical  Network for robust translation of image color temperature from input image to target image. These models have used a “coarse-to-fine” approach by exploiting multi-scale input images at different processing levels. Our end to end method is differentiable unlike other two step non-differentiable methods where color temperature transfer and shadow generation is taking place in two separate sub-networks with intermediate steps. Our model is also very lightweight and have a lower average inference time than other comparable multi-scale models. We have also demonstrated how our weighted combined loss function performs much better than differentiable sobel loss function which are used for getting reconstructed relighted images with more prominent edges. In addition, we have also shown that with two distinct loss functions, using a multi-step training approach to this problem can significantly increase performance and can achieve a high-quality reconstruction of a relighted image.

\section{Related Work}
Deep learning based methods have played a vital role in relighting real-life and synthetic scenarios with great efficiency. Gafton \emph{et al.} \cite{gafton20202d} proposed a GAN based Image translation methods like pix2pix \cite{isola2017image}. There is no additional knowledge requirement of the view geometry and parameters to generate proper lighting effects in this type of approach. 
Zhou \emph{et al.} \cite{zhou2019deep} proposed the use of Spherical Harmonics lighting parameters \cite{sengupta2018sfsnet} in the bottleneck layer of the U-Net hourglass network to achieve state-of-the-art results over the portrait relighting task.
Ren \emph{et al.} \cite{ren2015image} introduces a regression-based neural network for relighting real-world scenes from a small number of images. They described relighting as the product of two matrices, a light transport matrix generated from input images, and new lighting vectors.
Xu \emph{et al.} \cite{xu2018deep} proposed a deep learning-based framework to generate images under novel illumination using only five images captured under predefined directional lights. Their framework included a fully convolution neural network to predict the relighting function and input light direction. 

\section{Proposed Approach}
\begin{figure}[h]
    \centering
    \includegraphics[scale=0.32]{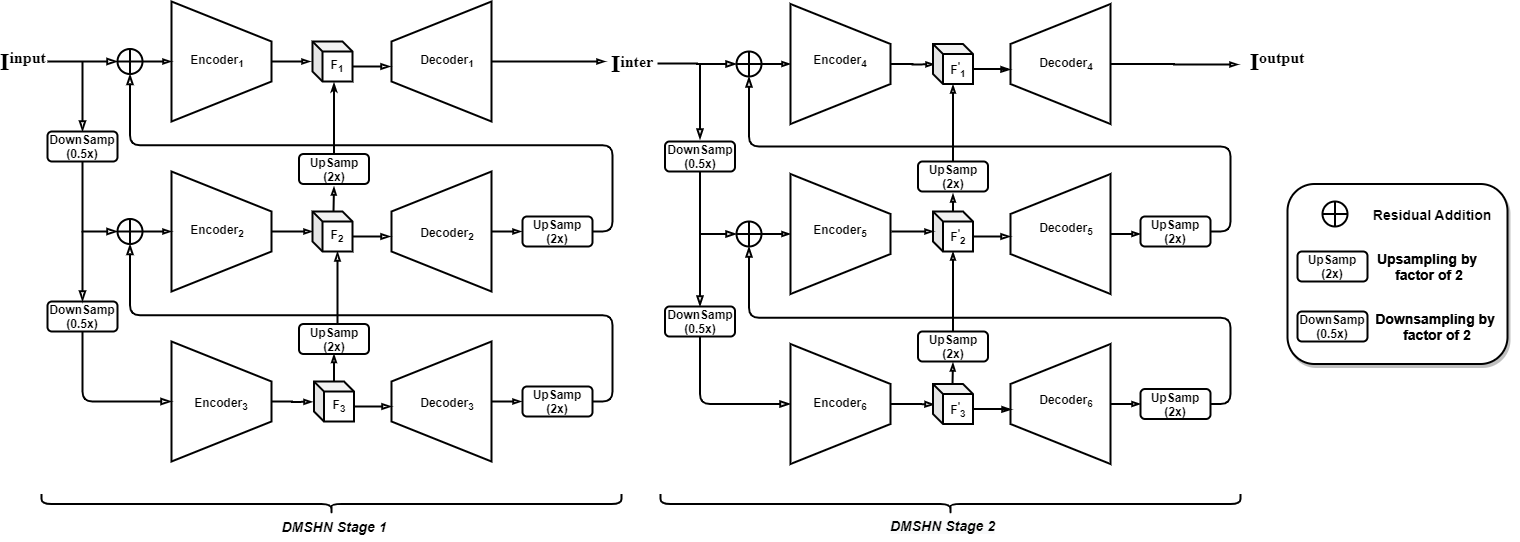}
    \caption{Architecture Diagram of Stacked DMSHN.}
    \label{fig:my_label}
\end{figure}

\textbf{Base Network:} We use a multi-scale network which works on a three-level image pyramid. In each level of the network, we have an encoder and a decoder. Decoder output of each level is upscaled and added to input image in next level and passed to encoder of next level. To capture global contextual information better, encoder output of each level is added with encoder output of previous level before feeding it to decoder. We call this base network Deep Multi-Scale Hierarchical Network (DMSHN) \cite{Das_fast_deep_2020}.  We can not improve performance of our model by adding additional lower level at DMSHN model as discussed in \cite{Zhang_2019_CVPR}. Thus, we cascade the same network twice to increase the performance of our network. We refer our final model as Stacked DMSHN. The model architecture diagram is shown in Fig. \ref{fig:my_label}. 

\textbf{Loss function:} We have done two stage training to enhance our results. First, we are using only $L_2$ loss as a reconstruction loss to train our model. Then, we used a linearly weighted loss function as objective loss function in second phase of training to get our final trained model. The combined loss function is given by,
\begin{equation}
    L_{CL} = \lambda_1*L_{1} + \lambda_2*L_{SSIM} + \lambda_3*L_p + \lambda_4*L_{tv}
\end{equation}
where, $L_{1}$ is Mean Absolute Error (MAE) loss, $L_{SSIM}$ is SSIM loss \cite{wang2004image} , $L_p$ is  Perceptual loss \cite{gatys2016image} and $L_{tv}$ is TV loss \cite{mahendran2015understanding}. During training, values of $\lambda_1$, $\lambda_2$, $\lambda_3$ and $\lambda_4$ are chosen to be $1$, $-5e-3$, $0.006$ and $2e-8$ respectively.

\section{Experiments}

\subsection{System Description}
We have implemented our code using Pytorch on a machine with  AMD 1950X, 64GB RAM and 11 GB NVIDIA GTX 1080 Ti GPU. We have used Adam optimizer to train our networks. During training, the Batch Size is 2 and we are resizing input image from $1024\times1024$ to $512\times512$. Initial learning rate is chosen to be $2e-3$ and the learning rate is gradually decreased to $5e-5$. We have trained our model with these settings for 2500 epochs. During testing, we feed the full resolution image to the network to get our output relighted image.

\subsection{Dataset Details}
We have used VIDIT dataset \cite{helou2020vidit}, a multi-scene synthetic dataset where the light source varies in 8 different equally-spaced azimuthal angles directions with a fixed polar angle, and five different colour temperatures. The dataset includes 390 different scenes that are captured with a total of 15,600 images at 40 different illumination settings for each illumination direction and colour temperature per scene. The dataset comprises of 390 samples with 300 images that have been used as a training dataset and 45 images being used as validation and testing each. In this paper, we have considered input illumination setting as light direction from North with color temperature 6500K and output illumination setting as light direction from East with color temperature 4500K in our experiments.

\subsection{Results}
\begin{figure*}[h]
    \centering
    \includegraphics[scale=0.38]{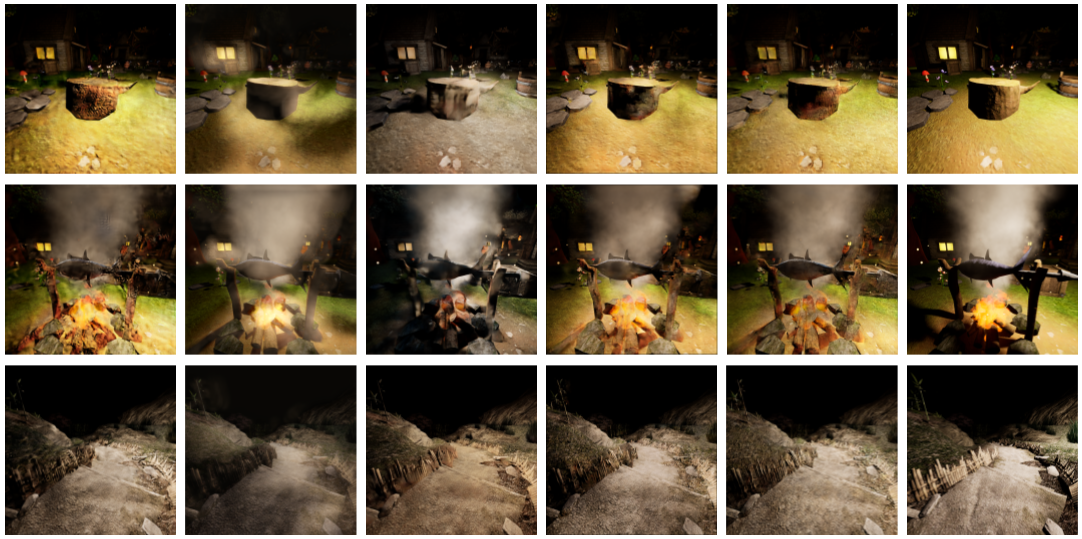}
    \caption{Qualitative Comparison of proposed method against Other Similar Multi-Scale Models. From left: (a) SRN \cite{tao2018scale} , (b) Dense-GridNet \cite{liuICCV2019GridDehazeNet}, (c) DRN\cite{wang2020deep}, (d) DMSHN (Ours) \cite{Das_fast_deep_2020}, (e) Stacked DMSHN (Ours), (f) Ground Truth.}
    \label{fig:other_model}
\end{figure*}

\textbf{Comparison with other methods:}  As ground truth for test set is not made publicly available, we are using validation set to show the qualitative and quantitative results for our model. Quantitative results on PSNR, SSIM \cite{wang2004image} and LPIPS \cite{zhang2018unreasonable} metrics are shown in the Table \ref{tab:singleview}. We have also compared our performance of the model with other two similar Multi-Scale models, Scale-Recurrent Network (SRN) \cite{tao2018scale} and Dense GridNet \cite{liuICCV2019GridDehazeNet}. We also compared our work with DRN\cite{wang2020deep} as they had worked on the same problem with same dataset. For qualitative comparison against other networks, example of output images are shown in Fig. \ref{fig:other_model}. From the both qualitative and quantitative results, it can be observed that two stage training of both DMSHN and Stacked DMSHN network significantly improved the performance of the model and in three out of four metrics, our method is achieving better performance than others. From the Fig. \ref{fig:other_model}, it can be inferred that predicted images from Dense-GridNet is dis-configured and have artifacts in spite of having the lowest LPIPS score but output images from our network have less artifacts and are perceptually better than the later one. Due to unavailability of depth information, our model has limited success in light gradient generation.

\begin{table*}[h]
\small
\centering
\tabcolsep=0.11cm
\begin{tabular}{c|c|c|c|c}
\hline
\textbf{Method} & \textbf{PSNR} & \textbf{SSIM} & \textbf{LPIPS} & \textbf{Runtime(s)} \\ \hline\hline
Dense-GridNet\cite{liuICCV2019GridDehazeNet} & 16.67 & 0.2811 & 0.3691 & 0.9326 \\ \hline
SRN\cite{tao2018scale} & 16.94 & 0.5660 & 0.4319 & 0.87 \\ \hline
DRN\cite{wang2020deep} & 17.59 & 0.596 & 0.440 & 0.5 \\ \hline
DMSHN\cite{Das_fast_deep_2020} (Our Method) & 17.20 & 0.5696 & 0.3712 & 0.0058 \\ \hline
Stacked DMSHN (Our Method) & 17.89 & 0.5899 & 0.4088 & 0.0116 \\ \hline
\end{tabular}
\caption{Quantitative Comparison of our method against Other Similar Multi-Scale Models on validation data.}
\label{tab_other_methods}
\end{table*}

\begin{figure*}[h]
    \centering
    \includegraphics[scale=0.38]{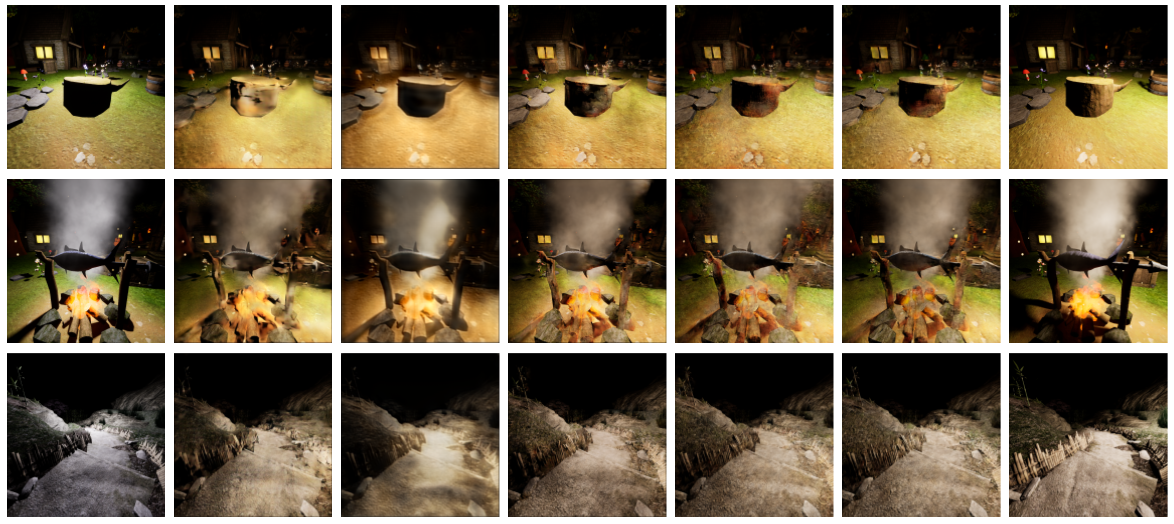}
    \caption{Single View Approach Results for both stage of training. From left: (a) Input Image, (b) DMSHN with Combined Sobel Loss, (c) DMSHN with $L_2$ Loss, (d) DMSHN with Combined Loss, (e) Stacked DMSHN with $L_2$ Loss, (f) Stacked DMSHN with Combined Loss, (g) Ground Truth.}
    \label{fig:singleview-results}
\end{figure*}

\textbf{Ablation Study:} We have experimented with different loss functions such as (a) A linear combination of $L_2$ loss and Advanced Sobel loss \cite{zheng2020image} $(L_{Sobel*} )$ as a Combined Sobel Loss ($L_{CSL}$) (b) Only $L_2$ loss $(L_{MSE} )$ and (c) Combined Loss $(L_{CL})$ to train our networks. In case (a) and (b), same loss function was used throughout the training, whereas in case (c) training was done in two stages. For DMSHN, $L_{CSL}$ produces best SSIM score, whereas $L_{CL}$ yields best scores for PSNR and LPIPS metrics as shown in Table \ref{tab:singleview}. $L_{CL}$ performs better than $L_{MSE}$ for Stacked DMSHN. Qualitative comparison for different model and loss functions are shown in Fig. \ref{fig:singleview-results}.

\begin{table*}[h]
\small
\centering
\tabcolsep=0.11cm
\begin{tabular}{c|c|c|c|c}
\hline
\textbf{Method} &\textbf{Loss Function} & \textbf{PSNR} & \textbf{SSIM} & \textbf{LPIPS} \\ \hline\hline
DMSHN & Combined Sobel Loss ($L_{CSL}$) & 17.09 & 0.5928 & 0.4428 \\ 
DMSHN & $L_2$ Loss ($L_{MSE}$) & 16.94 & 0.5659 & 0.4933 \\ 
DMSHN & Combined Loss ($L_{CL}$) & 17.20 & 0.5696 & 0.3712 \\ 
Stacked DMSHN & $L_2$ Loss ($L_{MSE}$) & 17.53 & 0.5673 & 0.4253 \\ 
Stacked DMSHN & Combined Loss ($L_{CL}$) & 17.89 & 0.5899 & 0.4088 \\
\hline

\end{tabular}
\caption{Quantitative Comparison of our method for different loss function and stages of training.}
\label{tab:singleview}
\end{table*}

\section{Conclusion}
In this paper, we have proposed a fast and efficient stacked network for one to one image relighting. It is a network with provisions for Multi-Scale training. Our model has the best runtime compared to currently available models to process a High resolution image. Our  method  is  fairly  generalized  and  can  be  utilized  for the tasks mentioned above. Though occluded dark parts of images cannot be generated by the network as there is no information present in that region to be recovered by the network. This problem can be mitigated to some extent if we have two input images illuminated from opposite directions, which can be a research direction to be explored in future. The network  is  fairly good for translating image color temperature from input image to target image and also performs moderately for light gradient generation with respect to target image.


\bibliographystyle{splncs}
\bibliography{egbib}





\end{document}